\pdfoutput=1

\documentclass[11pt]{article}

\usepackage[preprint]{acl}
\usepackage{times}
\usepackage{latexsym}
\usepackage[T1]{fontenc}
\usepackage[utf8]{inputenc}
\usepackage{colortbl}
\usepackage{xcolor}
\usepackage{tcolorbox}
\usepackage{microtype}
\usepackage{inconsolata}
\usepackage{graphicx}
\usepackage{subcaption}
\usepackage{booktabs}
\usepackage{multirow}
\usepackage{tablefootnote}
\usepackage{hyperref}
\tcbuselibrary{breakable} 
\usepackage{pifont}
\usepackage{amsmath} 
\usepackage{amssymb} 
\usepackage{amsfonts} 
\usepackage{algorithmic}
\usepackage{algorithm}
\definecolor{myblue}{RGB}{150, 150, 230}

\usepackage{amsmath,amsfonts,bm}

\def\eqref#1{equation~\ref{#1}}

\def\1{\bm{1}}

\DeclareMathAlphabet{\mathsfit}{\encodingdefault}{\sfdefault}{m}{sl}
\SetMathAlphabet{\mathsfit}{bold}{\encodingdefault}{\sfdefault}{bx}{n}

\title{
MAM: Modular Multi-Agent Framework for Multi-Modal Medical Diagnosis via Role-Specialized Collaboration
}

\author{
Yucheng Zhou, 
Lingran Song, 
Jianbing Shen\thanks{~Corresponding author. This work was supported by the National Natural Science Foundation of China (No. 624B2002) and the Jiangyin Hi-tech Industrial Development Zone under the Taihu Innovation Scheme (EF2025-00003-SKL-IOTSC).}\\
SKL-IOTSC, CIS, University of Macau \\
{\tt yucheng.zhou@connect.um.edu.mo, jianbingshen@um.edu.mo}
}

\begin{document}
\maketitle

\begin{abstract}
Recent advancements in medical Large Language Models (LLMs) have showcased their powerful reasoning and diagnostic capabilities. Despite their success, current unified multimodal medical LLMs face limitations in knowledge update costs, comprehensiveness, and flexibility.  
To address these challenges, we introduce the Modular Multi-Agent Framework for Multi-Modal Medical Diagnosis (MAM). 
Inspired by our empirical findings highlighting the benefits of role assignment and diagnostic discernment in LLMs, MAM decomposes the medical diagnostic process into specialized roles: a General Practitioner, Specialist Team, Radiologist, Medical Assistant, and Director, each embodied by an LLM-based agent. 
This modular and collaborative framework enables efficient knowledge updates and leverages existing medical LLMs and knowledge bases.  
Extensive experimental evaluations conducted on a wide range of publicly accessible multimodal medical datasets, incorporating text, image, audio, and video modalities, demonstrate that MAM consistently surpasses the performance of modality-specific LLMs. Notably, MAM achieves significant performance improvements ranging from 18\% to 365\% compared to baseline models.
Our code is released at \url{https://github.com/yczhou001/MAM}.
\end{abstract}

\section{Introduction}
Large Language Models (LLMs) have recently demonstrated remarkable reasoning capabilities ~\cite{radford2018improving,openai2023gpt4,DBLP:journals/corr/abs-2302-13971,DBLP:conf/aaai/YangZZZXJZ24,DBLP:conf/emnlp/ZhangCJYCCLWZXW23,DBLP:journals/corr/abs-2402-03300,DBLP:conf/emnlp/ZhangLB23}.
Beyond demonstrating impressive language reasoning and generation capabilities, LLMs are expanded to process diverse modalities, e.g., images, audio, and video~\cite{DBLP:conf/nips/LiuLWL23a,DBLP:journals/corr/abs-2311-07919,DBLP:conf/emnlp/ZhangLB23}.
This formidable reasoning capacity holds significant promise for addressing problems in medical diagnostics.

For medical practice, physicians are confronted with a deluge of heterogeneous medical data, encompassing textual reports, medical images, cardiac sounds, and even surgical video recordings.  
Accurately extracting critical information from this complex data to arrive at precise diagnoses places a significant cognitive burden and challenge on clinicians. 
Furthermore, the growth in the volume of medical diagnostic data provides a substantial foundation for the training of LLMs.
Consequently, the development of LLMs to enhance medical diagnostic workflows is crucial.

However, many efforts are directed towards constructing unified multimodal medical large models~\cite{llavamed,xrayGPT,ophglm}. 
While these models have shown some progress in integrating multimodal information, they suffer from two limitations.  
Firstly, for unified models, each knowledge update is cost, often requiring substantial computational resources to retrain the entire model.
Secondly, unified models lack modularity and flexibility, necessitating a single model to exhibit sufficient performance across various medical diagnostic tasks to satisfy demands.
To explore the capabilities of existing domain-specific LLMs, we conducted an empirical study.  Our findings indicate that role assignment significantly enhances the diagnostic abilities of LLMs, and LLMs possess the potential to discern the correct diagnosis from multiple ones.

To overcome the aforementioned limitations of unified multimodal medical LLMs and to better emulate the collaborative approach of human medical teams, we propose the Multi-Agent Framework for Multi-Modal Medical Diagnosis (MAM).  
Instead of pursuing an ``omnipotent'' unified model, MAM framework decomposes the medical diagnostic process into several specialized roles and designs LLM-based agents for each role. 
These agents include: a General Practitioner agent responsible for initial triage, a Specialist Team agent providing domain-expert opinions, a Radiologist agent specializing in medical image analysis, a Medical Assistant agent aiding information retrieval and knowledge management, and a Director agent responsible for coordinating and synthesizing diagnostic opinions. 
The core advantages of the MAM framework lie in its modular design and collaborative workflow. 
The modular agent design enables more granular and efficient knowledge updates and model maintenance, without requiring global retraining. 
The framework also allows MAM to easily integrate and leverage various existing medical models and specialized knowledge bases. 

In the experiments, we evaluate our MAM framework in multimodal medical diagnosis tasks through comprehensive experiments on several publicly available multimodal medical datasets. 
Experimental results demonstrate that the MAM framework consistently outperforms specific-modal LLMs across various medical datasets and data modalities. 
In addition, we conduct ablation studies, consistency analysis, and sensitivity analyses regarding the number of discussion rounds and roles to gain deeper insights into the roles of individual components and the operational mechanisms of the framework. 

\section{Related Work}\label{app:related}
\subsection{LLM-based Multi-Agent}
With the rapid advancement of LLMs, their application across various tasks has become increasingly widespread ~\cite{radford2018improving,openai2023gpt4,DBLP:journals/corr/abs-2302-13971,DBLP:conf/aaai/YangZZZXJZ24,DBLP:conf/emnlp/ZhangCJYCCLWZXW23,DBLP:journals/corr/abs-2402-03300,DBLP:conf/emnlp/ZhangLB23,ZhouS025,ZhouLWS24}. These LLMs, outstanding in natural language processing, have been widely adapted to different tasks~\cite{DBLP:conf/iclr/WangRZLLSZSZ024,DBLP:conf/dasfaa/YueLZSWXLSSCHW24,DBLP:conf/aaai/YangZZZXJZ24,zhou2024less,hu2025beyond}. However, for relatively complex tasks, the capabilities of a single LLM may not achieve the desired effect. In order to solve complex tasks beyond the function of a single LLM, LLM-based multi-agent systems are developed~\cite{DBLP:journals/corr/abs-2310-02170,DBLP:conf/aaai/Zhao0XLLH24,DBLP:journals/corr/abs-2306-03314}. In~\cite{wu2024mathchat}, the author investigates the effectiveness of using LLM-based Multi-Agent to solve mathematical problems through dialogue, and MathChat is proposed as a conversational problem-solving solution designed for mathematical problems. In software engineering, the MAGIS~\cite{DBLP:conf/nips/0003ZWZ0C24} framework enables the collaboration of various agents in the planning and coding process to solve GITHUB problems. In the field of finance, inspired by the organizational structure of effective investment firms in the real world, FinCon~\cite{DBLP:conf/nips/YuYLDJCCSCLXZSX24} is developed to accomplish a variety of financial tasks.

The emergence of these LLM-based Multi-Agent systems points to a common conclusion that LLM-based Multi-Agents are well adapted for reasoning~\cite{DBLP:conf/aaai/Zhao0XLLH24,DBLP:journals/corr/abs-2306-03314,wu2024mathchat,DBLP:conf/nips/0003ZWZ0C24}, decision making~\cite{DBLP:journals/corr/abs-2310-02170,DBLP:conf/nips/YuYLDJCCSCLXZSX24}, etc. This adaptability makes them essential in the medical domain where interdisciplinary knowledge and multi-step problem solving are required.

\subsection{Medical LLM}
Due to the remarkable performance of LLMs in different tasks ~\cite{DBLP:conf/emnlp/ZhangCJYCCLWZXW23,DBLP:journals/corr/abs-2402-03300,DBLP:conf/emnlp/ZhangLB23}, various medical LLMs have been developed to solve a wide range of medical problems~\cite{DBLP:journals/corr/abs-2308-14346,DBLP:conf/aaai/Zhao0XLLH24,DBLP:journals/corr/abs-2310-09089,DBLP:conf/aaai/FlemingLHJRTBGS24}. As a comprehensive solution, DISC-MedLLM~\cite{DBLP:journals/corr/abs-2308-14346} utilizes LLM to deliver accurate and realistic medical responses in end-to-end conversational healthcare services. As the first LLaMA based Chinese medical LLM, Zhongjing~\cite{DBLP:conf/aaai/Zhao0XLLH24} has implemented the  training pipeline from continuous pre-training, SFT, to Reinforcement Learning from Human Feedback (RLHF), where pre-training enhances medical knowledge and RLHF further improves instruction compliance and safety. Through a multi-stage training approach that combines domain-specific continuous pre-training (DCPT), SFT, and Direct Preference Optimization (DPO), Qilin-Med~\cite{DBLP:journals/corr/abs-2310-09089} shows substantial performance gains as a medical LLM.

Moreover, medical field is characterized by the presence of multimodal information, including diverse data types such as text, images, audios, etc. To make full use of these diverse data types, multimodal medical large models have been created~\cite{llavamed,xrayGPT,ophglm,qilinMedVL,DBLP:journals/corr/abs-2406-19280}. In LLaVA-Med~\cite{llavamed}, the authors propose a cost-effective way to train a visual language conversational assistant that can answer open-ended research questions on biomedical images. In the task area of radiology, XrayGPT~\cite{xrayGPT} was developed, which is a new conversational medical visual language model that can analyze and answer open-ended questions about chest radiographs. In the field of ophthalmology, OphGLM~\cite{ophglm} has built a large multimodal model of ophthalmology, contributing to the clinical application of ophthalmology.

\section{Empirical Study}\label{empirical}
\subsection{The Significance of Assigned Roles in Medical Diagnosis with LLMs}
\paragraph{Experimental Setting.}
We investigate the impact of assigned roles on the performance of Large Language Models (LLMs) in multimodal medical diagnosis.  Our experiments leverage a diverse collection of publicly available medical datasets, encompassing text, image, audio, and video modalities.  Specifically, we utilize the following datasets: Brain Tumor \cite{braintumor} (394 cases), DeepLesion \cite{DBLP:journals/corr/abs-1710-01766} (225 cases), Heartbeat \cite{pascal-chsc-2011} (461 cases), MedQA \cite{DBLP:journals/corr/abs-2009-13081} (200 cases), MedVidQA \cite{medvidqa} (284 cases), NIH Chest X-rays \cite{DBLP:conf/cvpr/WangPLLBS17} (215 cases), PathVQA \cite{DBLP:journals/corr/abs-2003-10286} (200 cases), PMC-VQA \cite{DBLP:journals/corr/abs-2305-10415} (200 cases), PubMedQA \cite{DBLP:conf/emnlp/JinDLCL19} (200 cases), and SoundDr \cite{SoundDr} (240 cases). We use these datasets to evaluate the capabilities of LLMs in handling multimodal medical diagnostic tasks.
We emploly Qwen-Audio-Chat \cite{DBLP:journals/corr/abs-2311-07919} for audio tasks, Medichat-Llama3-8B \cite{medichat_llama3_8b} for text tasks, HuatuoGPT-Vision-7B \cite{DBLP:journals/corr/abs-2406-19280} for image tasks, and VideoLLaMA2-7B \cite{DBLP:journals/corr/abs-2406-07476} for video tasks. LLMs can assign roles in input prompts.

\begin{table}[t]\small
\centering
\resizebox{\linewidth}{!}{
\setlength{\tabcolsep}{3.2pt}
\begin{tabular}{lcc}
\toprule
\bf Dataset & \bf Direct & \bf Assigned Roles\\ \midrule
MedQA  \cite{DBLP:journals/corr/abs-2009-13081}  & 30.8 & 50.6 \textcolor{blue}{(+19.8)}\\
PubMedQA \cite{DBLP:conf/emnlp/JinDLCL19} & 48.5  & 87.0 \textcolor{blue}{(+38.5)}\\
PathVQA \cite{DBLP:journals/corr/abs-2003-10286} & 40.1  & 46.6 \textcolor{blue}{(+6.5)}~~~\\
PMC-VQA \cite{DBLP:journals/corr/abs-2305-10415} & 24.0  & 29.0 \textcolor{blue}{(+5.0)}~~~\\
DeepLesion \cite{DBLP:journals/corr/abs-1710-01766} & 11.1  & 40.0 \textcolor{blue}{(+28.9)}\\
NIH \cite{DBLP:conf/cvpr/WangPLLBS17} & 12.6  & 50.7 \textcolor{blue}{(+38.1)}\\
Brain Tumor \cite{braintumor} & 80.2  & 98.2 \textcolor{blue}{(+18.0)}\\
Heartbeat \cite{pascal-chsc-2011} & 43.9  & 62.5 \textcolor{blue}{(+18.6)}\\
SoundDr \cite{SoundDr} & 25.0  & 45.4 \textcolor{blue}{(+20.4)}\\
MedVidQA \cite{medvidqa}  &  55.3 & 69.7 \textcolor{blue}{(+14.4)}\\ \bottomrule
\end{tabular}}
\caption{\small Performance comparison of ``Direct'' and ``Assigned Roles'' prompting methods across multi-modal medical tasks. The {\color{blue} blue} numbers indicate the performance improvement.}
\label{tab:roles}
\end{table}

Table~\ref{tab:roles} compares two prompting strategies: ``Direct'' and ``Assigned Roles''. The ``Direct'' is without role assignment, while the ``Assigned Roles'' approach creates a physician role using a specific prompt (see Appendix~\ref{app:prompt}).
The results indicate a consistent and significant performance improvement across all datasets using the ``Assigned Roles'' prompting strategy, with gains ranging from 5.0\% (PMC-VQA) to 38.5\% (PubMedQA). This suggests that role context enhances LLMs' ability to interpret and reason about medical data, improving diagnostic accuracy. Even in datasets with high baseline performance (e.g., Brain Tumor), role assignment provides a noticeable benefit. Furthermore, role assignment is particularly effective for tasks requiring deeper medical context and domain-specific knowledge.

\subsection{LLMs' Capability to Discern Correct Reasoning Outcomes}
\begin{table}[t]\small
\centering
\resizebox{\linewidth}{!}{
\setlength{\tabcolsep}{2pt}
\begin{tabular}{lcc}
\toprule
\bf Dataset & \bf Expectation & \bf Reasoning\\ \midrule
MedQA \cite{DBLP:journals/corr/abs-2009-13081}  & 46.9 & 49.3 \textcolor{blue}{(+2.4)}~~~\\
PubMedQA \cite{DBLP:conf/emnlp/JinDLCL19} &  54.0 & 72.6 \textcolor{blue}{(+18.6)}\\
PathVQA \cite{DBLP:journals/corr/abs-2003-10286} &  50.0 & 91.7 \textcolor{blue}{(+41.7)}\\
PMC-VQA \cite{DBLP:journals/corr/abs-2305-10415} & 41.7  & 62.5 \textcolor{blue}{(+20.8)}\\
DeepLesion \cite{DBLP:journals/corr/abs-1710-01766} & 52.5  & 57.5 \textcolor{blue}{(+5.0)}~~~\\
NIH \cite{DBLP:conf/cvpr/WangPLLBS17} &  44.0 & 46.0 \textcolor{blue}{(+2.0)}~~~\\
Brain Tumor \cite{braintumor} & 55.1  & 73.9 \textcolor{blue}{(+18.8)}\\
Heartbeat \cite{pascal-chsc-2011} &  50.2 & 54.0 \textcolor{blue}{(+3.8)}~~~\\
SoundDr \cite{SoundDr} & 48.1  & 55.7 \textcolor{blue}{(+7.6)}~~~\\
MedVidQA \cite{medvidqa}  &  46.0 & 50.0 \textcolor{blue}{(+4.0)}~~~\\ \bottomrule
\end{tabular}}
\caption{\small LLM Diagnostic Discernment: Comparing Expected vs. Reasoning Accuracy. This table shows the performance of LLMs in selecting the correct diagnosis from a set of plausible alternatives. ``Expectation'' represents random selection accuracy, while ``Reasoning'' reflects the LLM's accuracy in identifying the correct diagnosis.}
\label{tab:discern}
\end{table}

\begin{figure*}[t]
\centering
\includegraphics[width=1\linewidth]{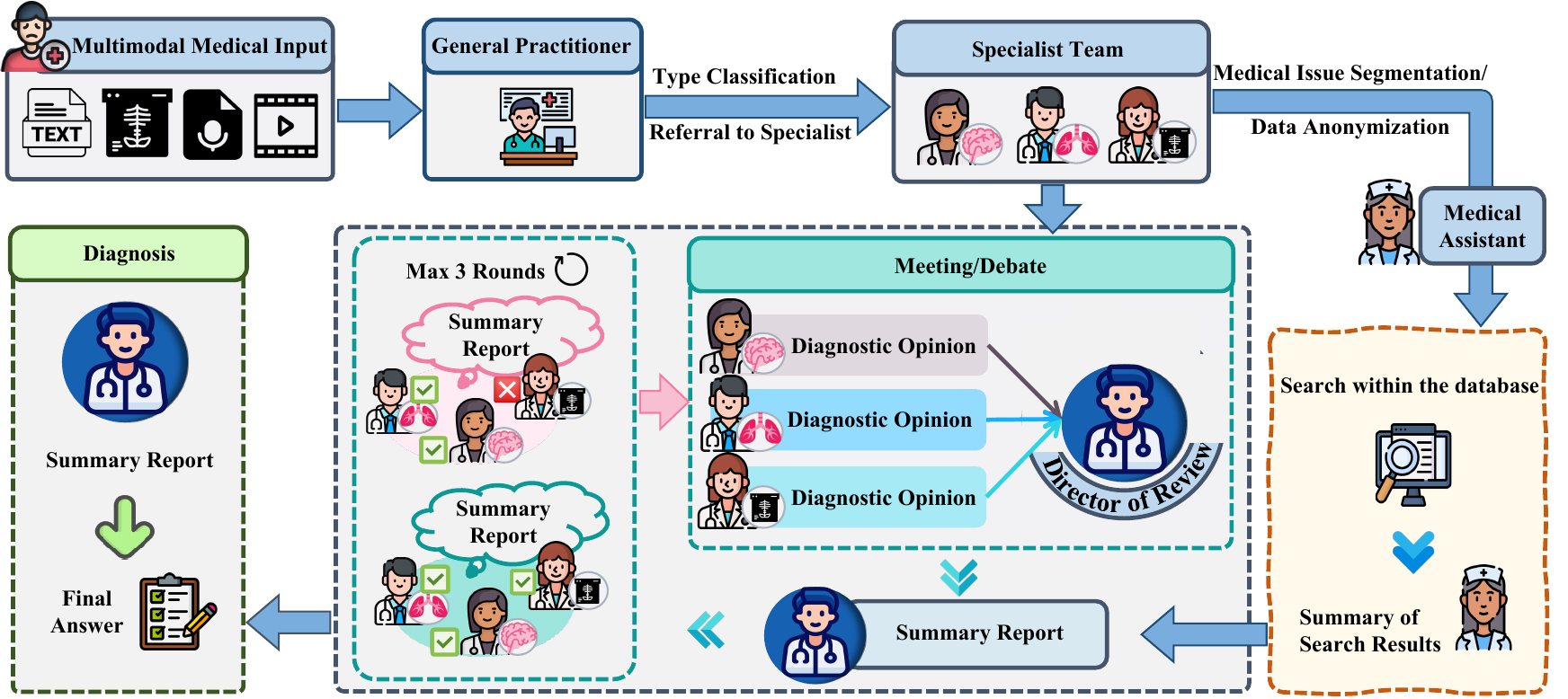}
\caption{\small Overview of our MAM Framework for multi-modal medical diagnosis.}
\label{fig:method}
\end{figure*}

To evaluate the reasoning capabilities of LLMs, we designed an experiment to assess their ability to identify the correct diagnosis from a set of plausible alternatives. Using the ``Assigned Roles'' prompting strategy, we generated three diagnostic outputs for each instance in the datasets. The datasets were pre-filtered to include only instances where at least one of the three diagnoses was correct, establishing a baseline ``Expectation'' representing random selection accuracy. The ``Reasoning'' column in Table~\ref{tab:discern} reflects the LLMs' accuracy in explicitly selecting the correct diagnosis from three generated options.
Results show that the ``Reasoning'' accuracy consistently exceeds the ``Expectation'' across all datasets, with improvements ranging from 2.0\% (NIH) to 41.7\% (PathVQA). This demonstrates that LLMs possess reasoning capabilities beyond random selection, particularly in complex visual question answering tasks like PathVQA and PMC-VQA. The consistent positive delta across datasets indicates LLMs' potential to evaluate and refine their outputs to identify accurate conclusions.

\section{Method}
Based on our preliminary empirical studies, we observed that augmenting Large Language Models (LLMs) with specific medical roles significantly enhances their diagnostic performance. Furthermore, LLMs demonstrate a notable capacity to reason and synthesize correct diagnoses from diverse diagnostic opinions.  Inspired by these findings, we propose the Multi-Agent Medical (MAM) framework, shown in Figure~\ref{fig:method}. This framework aims to transform multi-modal medical diagnosis into a collaborative endeavor, thereby amplifying the diagnostic capabilities of existing models. It comprises five key roles, each embodied by an LLM-based agent, working synergistically within a defined workflow: \ding{172} \textbf{General Practitioner}: Responsible for initial \textit{Disease Type Classification} and \textit{Referral to Specialist}. \ding{173} \textbf{Specialist Team}:  Charged with providing \textit{Diagnostic Opinions} on specific medical conditions and actively participating in discussions. \ding{174} \textbf{Radiologist}:  Tasked with analyzing medical images and contributing to diagnostic discussions. \ding{175} \textbf{Medical Assistant}:  Responsible for retrieving and summarizing relevant medical information from databases. \ding{176} \textbf{Director}:  Synthesizes discussion reports and reviews the quality of medical diagnoses.

In our framework, multi-modal medical inputs are initially directed to the General Practitioner, who performs disease classification and subsequently refers the case to the relevant Specialist Team. The Specialist Teams will decompose and anonymize the medical problem. The Medical Assistant then retrieves and summarizes pertinent information from medical databases based on the decomposed problem. Subsequently, the Director orchestrates discussions among the Specialist Team, where each specialist presents their diagnostic opinion.  The Director then synthesizes these opinions and the database summaries into a comprehensive report. The Specialist Team reviews this report and votes on whether to endorse it. In cases of disagreement, the process iteratively re-enters the Specialist Team discussion phase. Otherwise, upon reaching a consensus, the Director derives the final diagnosis based on the synthesized report.

\subsection{Doctor Agent Role Design}
\paragraph{\raisebox{-.3\baselineskip}{\includegraphics[height=1.1\baselineskip]{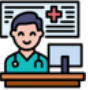}}~\textbf{General Practitioner}}
The General Practitioner agent is designed to mimic the role of a primary care physician in a clinical setting.  Upon receiving multi-modal medical inputs, this agent is responsible for the initial triage, performing \textit{Disease Type Classification} to categorize the medical case. Crucially, it then determines the appropriate \textit{Referral to Specialist}, directing cases to the relevant Specialist Team based on the initial classification.

\paragraph{\raisebox{-.3\baselineskip}{\includegraphics[height=1.1\baselineskip]{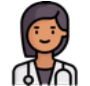}}~\textbf{Specialist Team}}
The Specialist Team is composed of multiple agents, each representing a specialist in a specific medical domain. These agents are tasked with providing \textit{Diagnostic Opinions} relevant to their expertise.  They engage in discussions, sharing their perspectives and interpretations of the medical case.  Furthermore, Specialist Team members participate in a voting process to reach a consensus on the synthesized diagnostic report.

\paragraph{\raisebox{-.3\baselineskip}{\includegraphics[height=1.1\baselineskip]{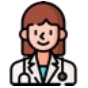}}~\textbf{Radiologist}}
The Radiologist agent specializes in the interpretation of medical images, such as X-rays and CT scans.  Its primary responsibility is to analyze these images and provide imaging-based insights to the other agents.  The Radiologist communicates with the Specialist Team and the Director, offering expertise in image analysis to aid in diagnosis and treatment planning.

\paragraph{\raisebox{-.3\baselineskip}{\includegraphics[height=1.1\baselineskip]{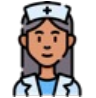}}~\textbf{Medical Assistant}}
The Medical Assistant agent plays a crucial role in information management.  Its responsibilities include processing medical data to facilitate retrieval of relevant information from medical databases.  Furthermore, the Medical Assistant summarizes the retrieved information, providing concise summaries.

\paragraph{\raisebox{-.3\baselineskip}{\includegraphics[height=1.1\baselineskip]{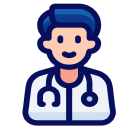}}~\textbf{Director}}
The Director agent serves as the orchestrator and synthesizer within the MAM framework.  This agent is responsible for reviewing the diagnostic opinions provided by the Specialist Team and the Radiologist.  It synthesizes the discussion outcomes from the Specialist Team into a comprehensive report. Crucially, the Director derives the final diagnosis based on the synthesized report, specifically when a consensus is reached among the Specialist Team through voting.

\subsection{Collaborative Diagnosis Process}
The MAM framework orchestrates a collaborative diagnostic process initiated upon receiving multi-modal medical inputs. Let $M = \{m_1, m_2, ..., m_k\}$ represent the multi-modal medical input, where $m_i$ denotes the $i$-th modality.

\paragraph{Initial Triage and Referral:} The General Practitioner agent ($G$) receives the multi-modal input $M$. $G$ performs \textit{Disease Type Classification} to categorize the medical case into a disease type $d$. This can be represented as:
\begin{align}
d = C^G(M)
\label{eq:disease_classification}
\end{align}
where $C^G$ denotes the disease type classification function performed by agent $G$.

Based on the classified disease type $d$, $G$ determines the appropriate Specialist Team $S = \{s_1, s_2, ..., s_n\}$ for referral. This referral process can be represented by:
\begin{align}
S = R^G(d)
\label{eq:referral}
\end{align}
where $R^G$ is the referral function by agent $G$, and $S$ is the set of specialist agents $s_i$.

\paragraph{Problem Decomposition and Anonymization:} The Specialist Team $S$ receives the medical case and decomposes the problem into a set of sub-problems $P = \{p_1, p_2, ..., p_m\}$. Anonymization is performed concurrently.  This decomposition is represented by:
\begin{align}
P = D^S(M)
\label{eq:decomposition}
\end{align}
where $D^S$ is the problem decomposition function by Specialist Team $S$.

\paragraph{Information Retrieval:} The Medical Assistant agent ($A$) utilizes the decomposed problem $P$ to retrieve relevant medical information. Let $I_r$ represent the retrieved information, obtained through:
\begin{align}
I_r = \text{Retrieve}^A(P)
\label{eq:retrieval}
\end{align}
where $\text{Retrieve}^A$ is the information retrieval function by agent $A$.
Since we do not have access to a real hospital database, the retrieval process is conducted using the Google API. The query used for retrieval is based on the decomposed and anonymized problem $P$, ensuring no privacy leakage. 
$A$ then summarizes the retrieved information into a concise summary $I_s$:
\begin{align}
I_s = \text{Summarize}^A(I_r)
\label{eq:summarization}
\end{align}
where $\text{Summarize}^A$ is the summarization function by agent $A$.

\paragraph{Diagnostic Opinion Generation and Discussion:} Each specialist $s_i \in S$ and the Radiologist agent ($Rad$) independently generate their diagnostic opinions based on the multi-modal input $M$, and the information summary $I_s$. Let $O_{s_i}$ be the diagnostic opinion of specialist $s_i$ and $O_{Rad}$ be the opinion of the Radiologist. Opinions are generated through:
\begin{align}
O_{s_i} &= \text{Diag}^{s_i}(M, I_s) \quad \forall s_i \in S \label{eq:specialist_opinion}\\
O_{Rad} &= \text{Diag}^{Rad}(M) \label{eq:radiologist_opinion}
\end{align}
where $\text{Diag}^{s_i}$ and $\text{Diag}^{Rad}$ are the diagnostic opinion generation functions for specialist $s_i$ and Radiologist $Rad$, respectively. The Director agent ($Dir$) orchestrates a discussion.

\paragraph{Report Synthesis and Review:} The Director agent ($Dir$) synthesizes the diagnostic opinions $\{O_{s_1}, O_{s_2}, ..., O_{s_n}, O_{Rad}\}$ and the information summary $I_s$ into a comprehensive diagnostic report $R_p$. This synthesis is performed by:
\begin{align}
R_p = \text{Synth}^{Dir}(\{O_{s_i}\}_{s_i \in S}, O_{Rad}, I_s)
\label{eq:report_synthesis}
\end{align}
where $\text{Synth}^{Dir}$ is the synthesis function by agent $Dir$. The Specialist Team $S$ reviews $R_p$ and votes on endorsement. Let $v_i \in \{0, 1\}$ be the vote of specialist $s_i$.

\paragraph{Consensus and Iteration:} The Director agent checks for consensus. Let $V = \sum_{i=1}^{n} v_i$ be the total endorsement votes. If $V = n$ (assuming $n=3$ in your algorithm description seems to be a typo, it should be consensus of all specialists which is $V=n$), consensus is reached.

\paragraph{Final Diagnosis Derivation:} Upon reaching consensus, the Director agent derives the final diagnosis $D_{final}$ based on $R_p$. This is performed by:
\begin{align}
D_{final} = \text{Diagnosis}^{Dir}(R_p)
\label{eq:final_diagnosis}
\end{align}
where $\text{Diagnosis}^{Dir}$ is the diagnosis derivation function by agent $Dir$. $D_{final}$ is the output of the MAM framework.

\begin{algorithm}[t]
\caption{Consensus and Iteration Process}
\label{alg:consensus}
\begin{algorithmic}[1]
\WHILE{No Consensus}
    \STATE Specialist Team $S$ and Radiologist $Rad$ present and discuss diagnostic opinions $\{O_{s_i}\}_{s_i \in S}$ and $O_{Rad}$.
    \STATE Director $Dir$ synthesizes a report $R_p = Synth_{Dir}(\{O_{s_i}\}_{s_i \in S}, O_{Rad}, I_s)$.
    \STATE Specialist Team $S$ reviews report $R_p$.
    \FOR{each specialist $s_i \in S$}
        \STATE Specialist $s_i$ votes $v_i \in \{0, 1\}$ on endorsing $R_p$.
    \ENDFOR
    \STATE Calculate total endorsement votes $V = \sum_{i=1}^{n} v_i$.
    \IF{$V == n$}
        \STATE Consensus Reached $\leftarrow$ True.
    \ELSE
        \STATE Consensus Reached $\leftarrow$ False.
    \ENDIF
\ENDWHILE
\STATE \textbf{return} Consensus Reached
\end{algorithmic}
\end{algorithm}

\section{Experiments}
\subsection{Setup}
\paragraph{Text-based Evaluation.}
For text-based evaluation, we used MedQA \cite{DBLP:journals/corr/abs-2009-13081} and PubMedQA \cite{DBLP:conf/emnlp/JinDLCL19}. We selected 200 English four-option multiple-choice questions from MedQA \cite{DBLP:journals/corr/abs-2009-13081} and 200 question-answer pairs from PubMedQA \cite{DBLP:conf/emnlp/JinDLCL19}.  We compared MAM against LLMs: LLaMA-7B \cite{DBLP:journals/corr/abs-2307-09288}, DAPT-7B \cite{DBLP:conf/acl/GururanganMSLBD20}, MedAlpaca-7B \cite{DBLP:journals/corr/abs-2304-08247}, AdaptLLM-7B \cite{DBLP:journals/corr/abs-2411-19930}, LLaMA-3-8B \cite{llama3modelcard}, and Medichat-Llama3-8B \cite{medichat_llama3_8b}.

\paragraph{Image-based Evaluation.}
For image evaluation, we used Brain Tumor \cite{braintumor} (test set, 394 cases), DeepLesion \cite{DBLP:journals/corr/abs-1710-01766} (225 cases from 9 categories), NIH Chest X-rays \cite{DBLP:conf/cvpr/WangPLLBS17} (215 cases), PathVQA \cite{DBLP:journals/corr/abs-2003-10286} (200 cases), and PMC-VQA \cite{DBLP:journals/corr/abs-2305-10415} (200 pairs). Compared LVLMs include LLaVA-7B \cite{DBLP:conf/nips/LiuLWL23a}, Qwen2-VL-7B \cite{DBLP:journals/corr/abs-2409-12191}, LLaVA-Med-7B \cite{llavamed}, Qilin-Med-VL-13B \cite{qilinMedVL}, and HuatuoGPT-Vision-7B \cite{DBLP:journals/corr/abs-2406-19280}.

\paragraph{Audio-based Evaluation.}
For audio evaluation, we used Heartbeat \cite{pascal-chsc-2011} (clinical trial data, 461 instances) and SoundDr \cite{SoundDr} (240 instances). MAM is compared with Qwen-Audio-Chat \cite{DBLP:journals/corr/abs-2311-07919}.

\paragraph{Video-based Evaluation.}
For video evaluation, we used MedVidQA (Temporal Segment Prediction test set, 284 data points \cite{medvidqa}). We preprocessed it into yes/no questions using original and alternative video segments. MAM is compared with video-LLMs: LLaVA-Next-Video-7B \cite{zhang2024llavanextvideo}, Qwen2-VL-7B \cite{DBLP:journals/corr/abs-2409-12191}, and VideoLLaMA2-7B \cite{DBLP:journals/corr/abs-2406-07476}.

\begin{table}[t]\small
\centering
\resizebox{\linewidth}{!}{
\setlength{\tabcolsep}{2pt}
\begin{tabular}{lcc}
\toprule
\bf Method & \bf MedQA & \bf PubMedQA \\
\midrule
LLaMA-7B \cite{DBLP:journals/corr/abs-2307-09288}  & 18.6 & 37.2 \\
DAPT-7B \cite{DBLP:conf/acl/GururanganMSLBD20}  & 25.7 & 44.1 \\
MedAlpaca-7B \cite{DBLP:journals/corr/abs-2304-08247}  & 29.3 & 51.2 \\ 
AdaptLLM-7B  \cite{DBLP:journals/corr/abs-2411-19930}  & 30.5 & 56.8 \\ 
LLaMA-3-8B \cite{llama3modelcard}  & 29.6 & 43.6 \\
Medichat-Llama3-8B \cite{medichat_llama3_8b} & 30.8 & 48.5 \\
MAM & \bf 40.0 & \bf 84.0 \\
\bottomrule
\end{tabular}}
\caption{\small Performance comparison of different LLMs on text-based medical datasets.}
\label{tab:text}
\end{table}

\begin{table}[t]\small
\centering
\resizebox{\linewidth}{!}{
\setlength{\tabcolsep}{2pt}
\begin{tabular}{lccccc}
\toprule
\bf Method & \bf Pa & \bf PMC & \bf DL & \bf NIH & \bf BT \\
\midrule
LLaVA-7B \cite{DBLP:conf/nips/LiuLWL23a}  & 7.3  & 6.2  & 2.6  & 4.2  & 34.6 \\
Qwen2-VL-7B \cite{DBLP:journals/corr/abs-2409-12191}  & 29.5 & 10.6 & 3.6  & 6.3  & 52.6 \\
LLaVA-Med-7B \cite{llavamed}  & 36.3 & 19.8 & 8.5  & 9.2  & 73.7 \\
Qilin-Med-VL-13B \cite{qilinMedVL}  & 39.2 & 22.5 & 11.3 & 11.4 & 80.6 \\
HuatuoGPT-Vision-7B \cite{DBLP:journals/corr/abs-2406-19280}  & 40.1 & 24.0 & 11.1 & 12.6 & 80.2 \\
MAM                 & \bf 47.6 & \bf 32.5 & \bf 35.1 & \bf 58.6 & \bf 97.9 \\
\bottomrule
\end{tabular}}
\caption{\small Performance comparison of different LVLMs on various image-based medical datasets.``Pa'', ``PMC'', ``BT'' and ``DL'' denote ``PathVQA'', ``PMC-VQA'', ``Brain Tumor'' and ``DeepLesion''.}
\label{tab:image}
\end{table}

\begin{table}[t]\small
\centering
\setlength{\tabcolsep}{2pt}
\begin{tabular}{lcc}
\toprule
\bf Method & \bf Heartbeat & \bf SoundDr \\
\midrule
Qwen-Audio-Chat \cite{DBLP:journals/corr/abs-2311-07919} & 34.9 & 25.0 \\
MAM & \bf 64.0 & \bf 47.9 \\
\bottomrule
\end{tabular}
\caption{\small Performance comparison of audio-LLMs on audio-based medical datasets.}
\label{tab:audio}
\end{table}

\begin{table}[t]\small
\centering
\setlength{\tabcolsep}{2pt}
\begin{tabular}{lc}
\toprule
\bf Method & \bf MedVidQA \\
\midrule
LLaVA-Next-Video-7B \cite{zhang2024llavanextvideo} & 51.5 \\
Qwen2-VL-7B \cite{DBLP:journals/corr/abs-2409-12191}         & 54.8 \\
VideoLLaMA2-7B \cite{DBLP:journals/corr/abs-2406-07476}     & 55.3 \\
MAM                 & \bf 74.3 \\
\bottomrule
\end{tabular}
\caption{\small Performance comparison of different video-LLMs on the Video-based medical dataset.}
\label{tab:video}
\end{table}

\subsection{Main Results}
Our comprehensive experiments across text, image, audio, and video medical data demonstrate the superior performance of the MAM framework. As shown in Table~\ref{tab:text}-\ref{tab:video}, MAM consistently outperforms strong competitors across all modalities and achieves significant performance improvements ranging from 18\% to 365\% compared to baseline models.
For text-based medical question answering (Table~\ref{tab:text}), MAM significantly surpasses baseline LLMs on MedQA and PubMedQA datasets, demonstrating enhanced medical text understanding. In image-based diagnosis (Table~\ref{tab:image}), MAM achieves top accuracy across PathVQA, PMC-VQA, DeepLesion, NIH Chest X-rays, and Brain Tumor datasets, with particularly substantial gains on DeepLesion and NIH Chest X-rays. 
Audio-based results on Heartbeat and SoundDr datasets (Table~\ref{tab:audio}) show MAM's clear advantage over audio-LLM baselines in medical audio interpretation. For video-based medical question answering on MedVidQA (Table~\ref{tab:video}), MAM achieves leading accuracy, outperforming all video-LLM competitors.
These results collectively demonstrate MAM's efficacy in multi-modal medical diagnosis, highlighting the benefits of its collaborative multi-agent approach. 

\subsection{Ablation Study}
\begin{table}[!t]\small
\centering
\setlength{\tabcolsep}{3pt}
\begin{tabular}{lcccc}
\toprule
\bf Dataset & \bf Direct & \bf +Roles & \bf +Discussion & \bf +Retrival \\ \hline\midrule
\multicolumn{5}{c}{\textit{Incrementally Added Function $\rightarrow$}} \\\midrule
MedQA & 30.8 & 31.0 & 32.5 & \bf 40.0\\
PubMedQA & 48.5 & 69.5 & 77.0 & \bf 84.0\\
PathVQA & 40.1 & 46.0 & 47.0 & \bf 47.6\\
PMC-VQA & 24.0 & 26.0 & 32.0 & \bf 32.5\\
DeepLesion & 11.1 & 33.8 & 34.7 & \bf 35.1\\
NIH & 12.6 & 36.0 & 38.6 & \bf 58.6\\
Brain Tumor & 80.2 & 92.4 & 97.0 & \bf 97.9\\
Heartbeat & 34.9 & 35.1 & 49.5 & \bf 64.0\\
SoundDr & 25.0 & 32.9 & 43.3 & \bf 47.9\\
MedVidQA & 55.3 & 58.0 & 60.6 & \bf 74.3\\
\bottomrule
\end{tabular}
\caption{\small Ablation study of our MAM framework. The ``Direct'' represents the baseline. From left to right, we incrementally add functions. ``+Retrieval'' is our full MAM framework. }
\label{tab:ablation}
\end{table}
To evaluate the contribution of each component in the MAM framework, we conducted an ablation study, with results shown in Table~\ref{tab:ablation}. The study systematically assesses the impact of incrementally adding key functionalities, starting from a baseline ``Direct'' approach (using the baseline LLM directly for diagnosis) and progressively integrating agent roles (+Roles), inter-agent discussion (+Discussion), and information retrieval (+Retrieval), representing the complete MAM framework.
The results reveal consistent performance improvements across all datasets as each component is added. The introduction of agent roles (+Roles) shows significant gains over the baseline, emphasizing the value of role specialization. Enabling discussion (+Discussion) further enhances performance, demonstrating the benefits of collaborative reasoning. Most notably, the full MAM framework (+Retrieval) achieves the highest performance, highlighting the synergistic effects of role specialization, collaborative discussion, and information retrieval. The substantial improvement from ``+Discussion'' to ``+Retrieval'' underscores the critical role of the Medical Assistant in enhancing diagnostic accuracy through relevant medical knowledge. These findings confirm the efficacy of each MAM component and their combined impact on multi-modal medical diagnosis.

\subsection{Consistency}
\begin{table}[t]\small
\centering
\setlength{\tabcolsep}{4pt}
\begin{tabular}{lcc}
\toprule
\bf Dataset & \bf Consistency & \bf MAM  \\ \midrule
MedQA \cite{DBLP:journals/corr/abs-2009-13081}  & \cellcolor{cyan!20}34.4 & \cellcolor{cyan!20}40.0 \\
PubMedQA  \cite{DBLP:conf/emnlp/JinDLCL19} & \cellcolor{cyan!40}74.2 & \cellcolor{cyan!40}84.0 \\
PathVQA \cite{DBLP:journals/corr/abs-2003-10286} & \cellcolor{cyan!20}50.0 & \cellcolor{cyan!20}47.6 \\
PMC-VQA \cite{DBLP:journals/corr/abs-2305-10415} & \cellcolor{cyan!5}14.6 & \cellcolor{cyan!5}32.5 \\
DeepLesion \cite{DBLP:journals/corr/abs-1710-01766} & \cellcolor{cyan!5}12.0 & \cellcolor{cyan!5}35.1\\
NIH \cite{DBLP:conf/cvpr/WangPLLBS17} & \cellcolor{cyan!20}59.3 & \cellcolor{cyan!20}58.6\\
Brain Tumor \cite{braintumor} & \cellcolor{cyan!40}97.5 & \cellcolor{cyan!40}97.9\\
Heartbeat \cite{pascal-chsc-2011} & \cellcolor{cyan!40}70.2 & \cellcolor{cyan!40}64.0 \\
SoundDr \cite{SoundDr} & \cellcolor{cyan!20}60.0 & \cellcolor{cyan!20}47.9\\
MedVidQA \cite{medvidqa} & \cellcolor{cyan!40}67.5 & \cellcolor{cyan!40}74.3\\ \bottomrule
\end{tabular}
\caption{\small Consistency of prediction results from baseline (Direct) and MAM. Rows with lighter \textcolor{cyan!90}{cyan} color indicate datasets where MAM has relatively lower performance.}
\label{tab:consistency_augmented}
\end{table}
To evaluate the MAM framework's behavior, we analyzed its prediction consistency compared to the ``Direct'' approach. 
Consistency is defined as the percentage of instances where MAM's final prediction aligns with a correct prediction from the ``Direct'' method. 
This metric assesses MAM's ability to retain and reinforce correct baseline predictions while correcting errors. Table~\ref{tab:consistency_augmented} compares the consistency scores with MAM's overall performance across datasets. 
Results indicate a positive correlation between MAM's performance and consistency. 
For instance, datasets, where MAM performs well, show high consistency, suggesting MAM effectively builds on the ``Direct'' method's correct predictions.
In contrast, datasets with lower performance, such as PMC-VQA and DeepLesion lower consistency. 
This implies that when the ``Direct'' achieves lower accuracy, MAM may introduce changes that slightly reduce consistency with the original correct predictions. 
Nevertheless, MAM generally outperforms the ``Direct'' method overall, as shown in Table~\ref{tab:ablation} and Table~\ref{tab:consistency_augmented}, indicating that its refinements enhance diagnostic accuracy despite occasional deviations from the baseline's correct predictions. This demonstrates that MAM actively improves predictions through its collaborative, knowledge-augmented framework rather than merely replicating the ``Direct'' approach.

\subsection{Discussion Time and Performance}
\begin{figure}[t]
\centering
\includegraphics[width=1\linewidth]{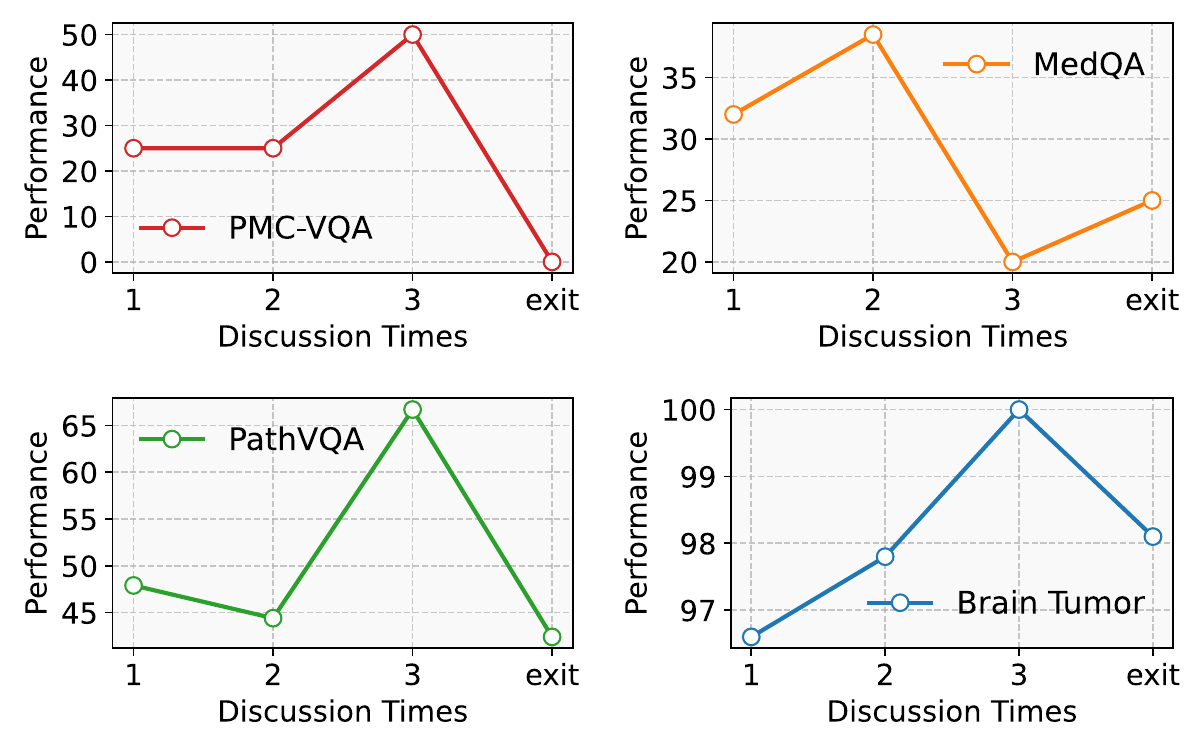}
\caption{\small Performance with different times of discussion ($\leq 3$) in our MAM pipeline across various datasets.}
\label{fig:discussion}
\end{figure}

We investigated the impact of iterative discussions on diagnostic accuracy by evaluating performance across different discussion rounds, as illustrated in Figure~\ref{fig:discussion}. For Brain Tumor, performance improved in early rounds, indicating that iterative discussions enhance accuracy for complex cases. However, extending discussions beyond a few rounds did not consistently yield further gains. For MedQA and PathVQA, performance fluctuated, with peak accuracy often achieved within the first two or three rounds. PMC-VQA experienced a performance decline in the final round, suggesting potential overfitting or dataset-specific issues. Results imply that while initial discussions can refine diagnoses, excessive rounds introduce noise or dilute accurate initial opinions. Limiting discussions can balance collaborative benefits and avoid over-discussion.

\subsection{Impact of Role Number}
\begin{figure}[t]
\centering
\includegraphics[width=1\linewidth]{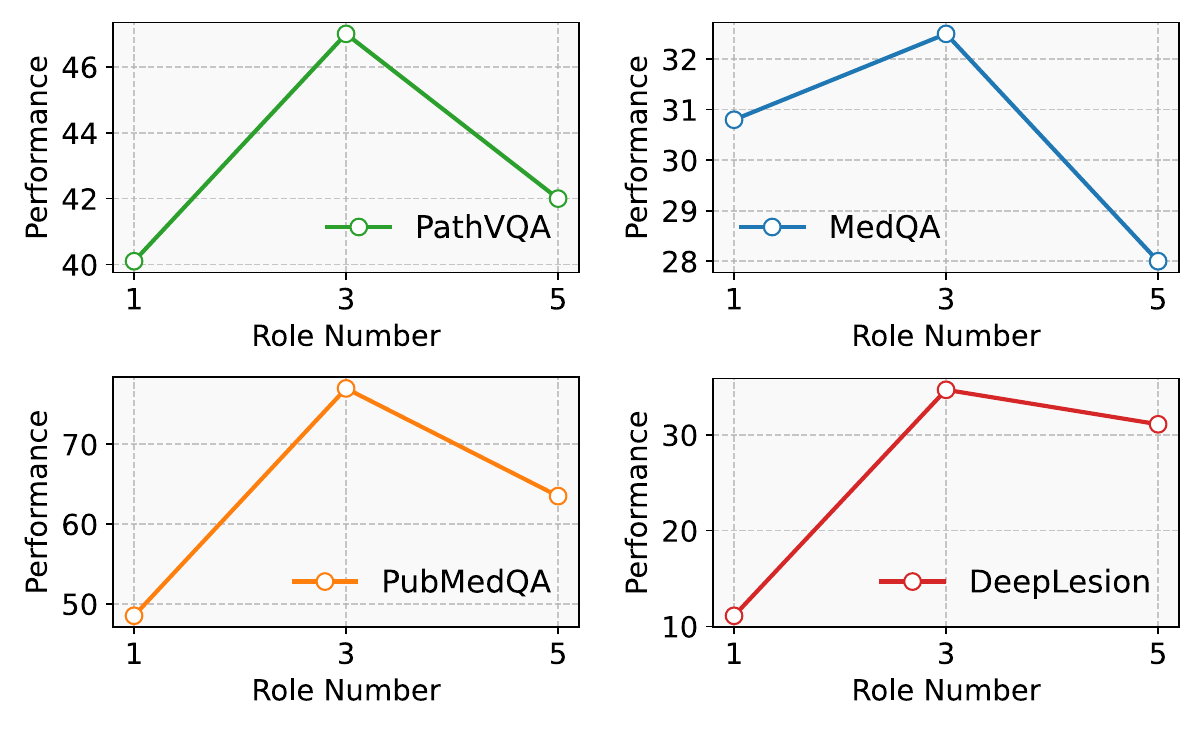}
\caption{\small Performance with different number of roles in our MAM pipeline across various datasets.}
\label{fig:RoleNumber}
\end{figure}

We investigate the effect of role granularity by varying the number of agents in MAM (Figure~\ref{fig:RoleNumber}).  Performance generally followed an inverted U-shape: increasing roles from 1 (``Direct'') to 3 significantly improved results, highlighting the benefit of role specialization. However, further increasing to 5 roles led to a performance decrease across datasets. This suggests an optimal level of role granularity exists. While role specialization is beneficial, excessive roles may introduce redundancy or overhead, hindering diagnosis.  A moderately specialized framework with 3 roles appears to strike a better balance than either a single-agent approach or an overly complex multi-agent system, indicating that streamlined role specialization is crucial for effective collaborative medical diagnosis.

\begin{table}[t]\small
\centering
\resizebox{\linewidth}{!}{
\setlength{\tabcolsep}{3pt}
\begin{tabular}{lcc}
\toprule
\bf Dataset & \bf Recall & \bf Answer Correct\\ \midrule
DeepLesion \cite{DBLP:journals/corr/abs-1710-01766}  & 31.7 & 53.4\\
Heartbeat \cite{pascal-chsc-2011}  & 34.0 & 58.8\\
NIH \cite{DBLP:conf/cvpr/WangPLLBS17}  & 12.1 & 46.2\\
\bottomrule
\end{tabular}}
\caption{\small ``Recall'' indicates the proportion of instances where the retrieved content includes the correct answer. ``Answer Correct'' represents the accuracy of the final answer under the retrieved content that encompasses the correct answer.}
\label{tab:recall}
\end{table}
\subsection{Recall of Retrieval}
To evaluate the Medical Assistant's information retrieval module, we first measured recall, defined as the proportion of retrieved medical documents containing information necessary to correctly answer diagnostic questions. As shown in Table~\ref{tab:recall}, recall varies across datasets, ranging from 12.1\% for NIH to 34.0\% for Heartbeat. These results indicate that while the module retrieves relevant information in some cases, significant improvement is needed. Imperfect recall may stem from limitations in retrieval algorithms, incomplete medical databases, or challenges in formulating effective search queries for diverse medical questions. Enhancing recall is critical for ensuring the availability of necessary information for downstream diagnostic tasks.

\subsection{Impact from Retrieval Content}
Complementary to evaluating retrieval recall, we examined the impact of retrieved content on diagnostic accuracy by calculating the conditional probability of obtaining a correct answer when the retrieved documents contained the necessary information. This metric, labeled ``Answer Correct'' in Table~\ref{tab:recall}, assesses the MAM framework's ability to leverage retrieved information effectively. As shown in Table~\ref{tab:recall}, the ``Answer Correct'' is consistently higher than the corresponding accuracy of the ``+Discussion'' (without retrieval) in Table~\ref{tab:ablation}. For instance, in the NIH dataset, the ``Answer Correct'' (46.2\%) significantly surpasses the ``Discussion'' (38.6\%). It demonstrates that when relevant information is retrieved, the MAM framework is more likely to arrive at a correct diagnosis. However, the ``Answer Correct'' is not perfect, demonstrating the importance of improving retrieval and LLM's reasoning capabilities.

\subsection{Case Study}
Figure \ref{fig:case} shows a case of multimodal input from DeepLesion, comparing the outputs from the baseline and our framework. The baseline model produced incorrect results, whereas our framework delivered correct predictions.
Our framework begins by identifying the input modality and determining the data type. Based on this information, it generates three expert roles to engage in up to three rounds of dialogue to discuss potential solutions. Concurrently, the Medical Assistant formulates queries for web retrieval. After processing the retrieved data, the Director reviews the discussion and retrieval records, synthesizes the insights into a summary, and presents it to the expert team for voting. The Director then uses the voting results and summary to make the final diagnosis.
As shown in the process discussed in Figure \ref{fig:case}, it is evident that although not all expert roles prioritized the correct answer initially, the structured approach of discussions and voting leads to an accurate resolution, which demonstrates effectiveness of our framework, demonstrating its ability of decision-making in complex medical scenarios.

\section{Conclusion}
This study introduces the Multi-Agent Framework for Multi-Modal Medical Diagnosis (MAM), addressing the limitations of unified multimodal medical LLMs. MAM employs a modular, collaborative approach, assigning specialized roles, i.e., General Practitioner, Specialist Team, Radiologist, Medical Assistant, and Director, to distinct LLM-based agents. This structure enhances knowledge updates, leverages specialized expertise, and adapts to diverse medical tasks and modalities. Extensive evaluations on multimodal medical datasets demonstrate MAM's superiority, outperforming modality-specific LLMs by 18\% to 365\%. Future work will integrate advanced knowledge retrieval and evaluate MAM in real-world clinical settings.

\section*{Limitations}
The performance of MAM is fundamentally constrained by the capabilities of the underlying LLMs utilized for each agent role. Inherent limitations such as model biases, knowledge gaps, or reasoning inaccuracies within these LLMs may propagate through the framework, potentially compromising diagnostic outcomes. MAM's architecture allows for flexible switching of base models, which could mitigate some limitations in future applications.
The other limitation of the current study is the absence of real-world clinical validation, which presents substantial challenges in terms of resource allocation and human expertise required for comprehensive evaluation. We acknowledge this limitation and propose to address it through clinical validation studies in our future work.

\bibliography{custom}
\clearpage
\appendix
\section{Case Study}\label{app:case}
\begin{figure*}[t]
\centering
\includegraphics[width=0.83\linewidth]{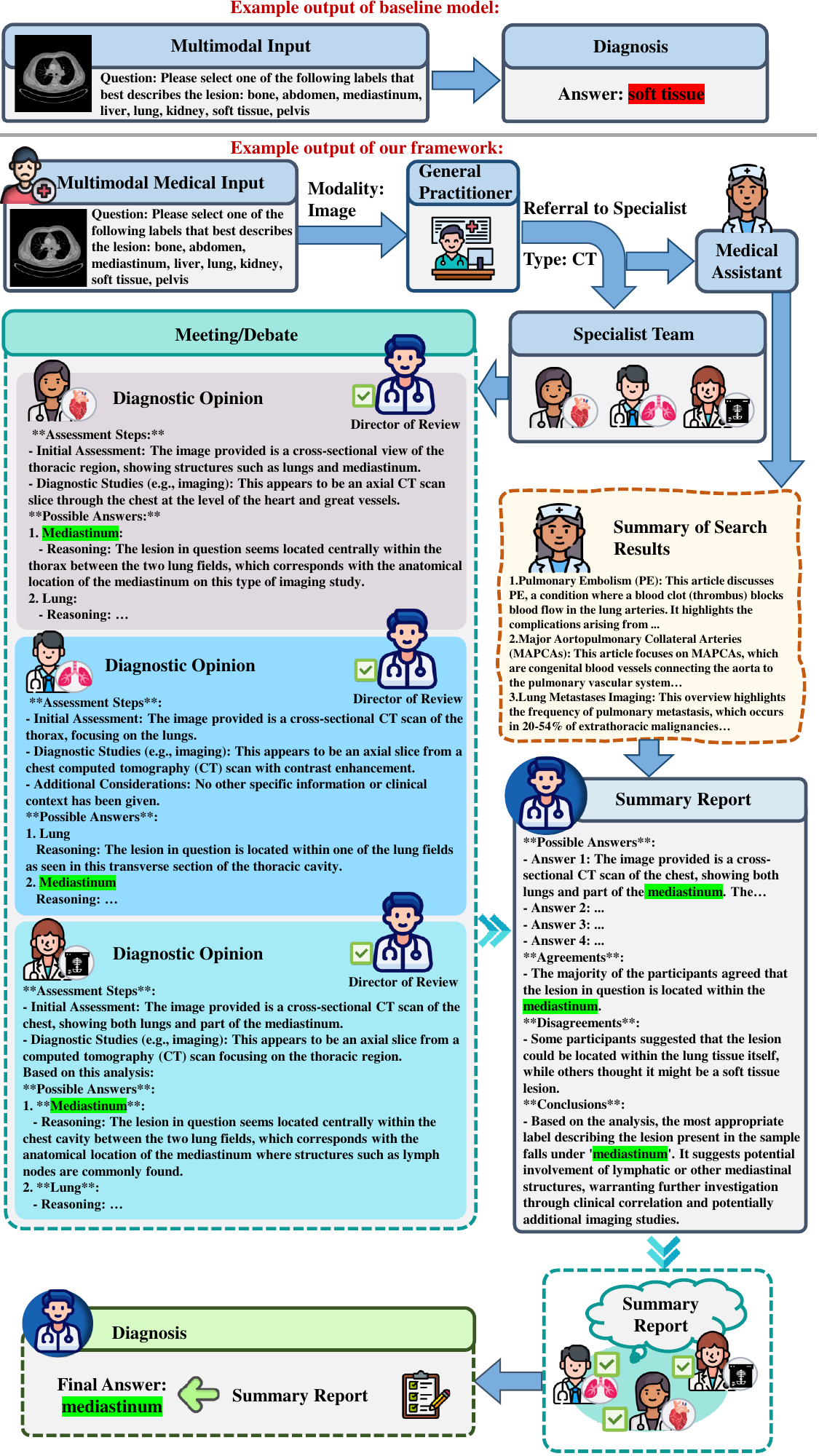}
\caption{\small Case Study.}
\label{fig:case}
\end{figure*}

Figure \ref{fig:case} shows the comparison results between the baseline model and our framework under the same sample input from DeepLesion \cite{DBLP:journals/corr/abs-1710-01766} dataset.

\section{Prompt}\label{app:prompt}

\begin{tcolorbox}[
colback=myblue!5!white,
colframe=myblue!75!black,
arc=1mm, 
auto outer arc,
title={Image Type Classification Prompt},
breakable
]\small

1.Please answer with a single word: What kind of medical image is this? X-Ray, CT, MRI, Pathology, Biomedical.

2.Please answer with a single word: What part of the human body does this image show? Brain, bone, abdomen, mediastinum, liver, lung, kidney, soft tissue, pelvis.

\end{tcolorbox}

\begin{tcolorbox}[
colback=myblue!5!white,
colframe=myblue!75!black,
arc=1mm, 
auto outer arc,
title={Audio Type Classification Prompt},
breakable
]\small
Please answer with a single word: What kind of audio is this? Cardiovascular, Respiratory.
\end{tcolorbox}

\begin{tcolorbox}[
colback=myblue!5!white,
colframe=myblue!75!black,
arc=1mm, 
auto outer arc,
title={Video Type Classification Prompt},
breakable
]\small
Please answer with a single word: What kind of video is this? Sports, Rehabilitation, Emergency.
\end{tcolorbox}

\begin{tcolorbox}[
colback=myblue!5!white,
colframe=myblue!75!black,
arc=1mm, 
auto outer arc,
title={Text Type Classification Prompt},
breakable
]\small
System prompt: You are given a question, please select a question type according to the given question.
        
Input: The question is $\{question\_text\}$. Which kind of question is this? Anaesthesia, Anatomy, Biochemistry, Dental, ENT, FM, O\&G, Medicine, Microbiology, Ophthalmology, Orthopaedics, Pathology, Pediatrics, Pharmacology, Physiology, Psychiatry, Radiology, Skin, PSM, Surgery, Unknown.
        
Output example: 

The question type is **Anaesthesia**.
\end{tcolorbox}

\begin{tcolorbox}[
colback=myblue!5!white,
colframe=myblue!75!black,
arc=1mm, 
auto outer arc,
title={Role Generation Prompt},
breakable
]\small

Given a disease type, generate a system prompt that assigns tasks to relevant medical roles, including **Specialist Doctor**, **Radiologic Technologist**, etc, from the perspective of a General Practitioner.\\
Input: The modality type is $\{modality\_type\}$, the disease type is $\{disease\_type\}$, and the patient question is $\{question\}$.\\
Output:\\
A system prompt that:

Identifies the relevant Specialist Doctor(s), Radiologic Technologist(s), and other Specialist(s) for the given disease type.

Assigns tasks to each identified role, specifying the necessary actions, tests, or examinations required for diagnosis and treatment.

Output example:

**Specialist Doctor** (Pulmonologist):

- Assess Patient's Health: Evaluate patient's function and overall health.

- Use $\{modality\_type\}$ Studies: Utilize expertise in the $\{disease\_type\}$ domains to diagnose diseases.

- Analyze Patient History and Symptoms: Determine the cause and severity of diseases by analyzing patient's medical history and symptoms.

\end{tcolorbox}

\begin{tcolorbox}[
colback=myblue!5!white,
colframe=myblue!75!black,
arc=1mm, 
auto outer arc,
title={Get Discuss Prompt},
breakable
]\small

You are a $\{role\_name\}$, responsible for the following tasks: $\{role\_responsibilities\}$. Please thoughtfully express your views for the following question.

Input: 

**Question type**: $\{disease\_type\}$.

**Question**: $\{question\}$.
    
Example output: 
    
**Assessment Steps**:

- Initial Assessment: [Provide a detailed overview of the initial assessment process]

- Diagnostic Studies (e.g., imaging, lab tests): [Include relevant details about any studies conducted]

- Additional Considerations: [Mention any other pertinent factors or evaluations]

**Possible Answers**:

- Answer 1: [Briefly explain answer 1]

Reasoning: [Briefly describe the corresponding reason for answer 1]

- Answer 2: [Briefly explain answer 2]

Reasoning: [Briefly describe the corresponding reason for answer 2]

- Answer 3: [Briefly explain answer 3]

Reasoning: [Briefly describe the corresponding reason for answer 3]

**Conclusion**: [Summarize the findings and provide a final recommendation or insight]

\end{tcolorbox}

\begin{tcolorbox}[
colback=myblue!5!white,
colframe=myblue!75!black,
arc=1mm, 
auto outer arc,
title={Get Summarize Prompt},
breakable
]\small

You are a specialized doctor serving as the moderator of this meeting. Please provide a detailed summary of the discussions that have taken place.
    
Example output:

**Possible Answers**:

- Answer 1: [Briefly explain answer 1]

- Answer 2: [Briefly explain answer 2]

- Answer 3: [Briefly explain answer 3]

**Agreements**:

- [Description of any agreements reached]

**Disagreements**:

- [Description of any disagreements that were noted]

**Conclusions**:

- [Final thoughts or conclusions drawn from the discussion]
    
Input: The question is $\{question\}$. The previous discussion of the meeting includes: $\{discussion\}$.

\end{tcolorbox}

\begin{tcolorbox}[
colback=myblue!5!white,
colframe=myblue!75!black,
arc=1mm, 
auto outer arc,
title={Get Vote Prompt},
breakable
]\small

You are a $\{role\_name\}$, responsible for the following tasks: $\{role\_responsibilities\}$. 

Please answer just using "yes" or "no" according to the following questions and the corresponding summery and the contents of the given file(if any).

Input: The question is $\{question\}$, and the summery of the discussion is:

$\{summary\}$

Do you agree with the summery above? Please answer just using "yes" or "no".

\end{tcolorbox}

\begin{tcolorbox}[
colback=myblue!5!white,
colframe=myblue!75!black,
arc=1mm, 
auto outer arc,
title={Get Review Prompt},
breakable
]\small

Question: Is there any medical reasoning errors, redundant statements, or invalid outputs in the following paragraph?

Please answer just using "yes" or "no".

Please read the rollowing paragraph:

$\{dis\}$

\end{tcolorbox}

\begin{tcolorbox}[
colback=myblue!5!white,
colframe=myblue!75!black,
arc=1mm, 
auto outer arc,
title={Get Multimodal Description Prompt},
breakable
]\small

Please describe this $\{modality\_type\}$ briefly in $100$ words:

\end{tcolorbox}

\begin{tcolorbox}[
colback=myblue!5!white,
colframe=myblue!75!black,
arc=1mm, 
auto outer arc,
title={Get Search Summarize Prompt},
breakable
]\small

Please summarize the following search results briefly in $200$ words:
$\{search\_result\}$

\end{tcolorbox}

\begin{tcolorbox}[
colback=myblue!5!white,
colframe=myblue!75!black,
arc=1mm, 
auto outer arc,
title={Get Diagnosis Prompt},
breakable
]\small

Input: Based on the provided image/video/audio (if applicable) and the meeting record, please provide answer to the following question.

Question: $\{ques\}$.

Meeting record: $\{record\}$.

\end{tcolorbox}

\begin{tcolorbox}[
colback=myblue!5!white,
colframe=myblue!75!black,
arc=1mm, 
auto outer arc,
title={Get Overall Review Prompt},
breakable
]\small

Input: You're a medical assistant. Please check whether the answer to this question is reasonable, if it is, please answer "yes", if not, please answer "no".

Question: $\{ques\}$.

Answer: $\{record\}$.

\end{tcolorbox}

\end{document}